\documentclass[11pt,a4paper]{article}
\usepackage[hyperref]{acl2020}
\usepackage{times}
\usepackage{amsmath}
\usepackage{amssymb}
\usepackage{graphicx}
\usepackage{latexsym}
\usepackage{subfigure}
\usepackage{longtable}
\usepackage{booktabs}
\usepackage{CJKutf8}

\usepackage{microtype}

\aclfinalcopy 

\title{{C}onversational {W}ord {E}mbedding for {R}etrieval-{B}ased {D}ialog {S}ystem}

\author{
Wentao Ma$^\dag$,
Yiming Cui$^\ddag$$^\dag$,
Ting Liu$^\ddag$,
Dong Wang$^\dag$,
Shijin Wang$^\dag$$^\S$,
Guoping Hu$^\dag$\\
{$^\dag$State Key Laboratory of Cognitive Intelligence, iFLYTEK Research, China} \\
{$^\ddag$Research Center for Social Computing and Information Retrieval (SCIR), } \\
{Harbin Institute of Technology, Harbin, China} \\
{$^\S$iFLYTEK AI Research (Hebei), Langfang, China} \\
$^\dag$$^\S$\tt\{wtma,ymcui,dongwang4,sjwang3,gphu\}@iflytek.com \\
$^\ddag$\tt\{ymcui,tliu\}@ir.hit.edu.cn
}

\date{}

\begin{document}
\maketitle
\begin{abstract}
 Human conversations contain many types of information, e.g., knowledge, common sense, and language habits.
 In this paper, we propose a conversational word embedding method named PR-Embedding, which utilizes the conversation pairs  $ \left\langle{post, reply} \right\rangle$  \footnote{In this paper, we name the first utterance in the conversation pair as `post,' and the latter is `reply'}  to learn word embedding. 
Different from previous works, PR-Embedding uses the vectors from two different semantic spaces to represent the words in post and reply.
To catch the information among the pair, we first introduce the word alignment model from statistical machine translation to generate the cross-sentence window, then train the embedding on word-level and sentence-level.
We evaluate the method on single-turn and multi-turn response selection tasks for retrieval-based dialog systems.
The experiment results show that PR-Embedding can improve the quality of the selected response. \footnote{PR-Embedding source code is available at \url {https://github.com/wtma/PR-Embedding} .}
\end{abstract}

\begin {figure*} [!t]
  \centering \small
  \includegraphics [width= 0.8\textwidth] {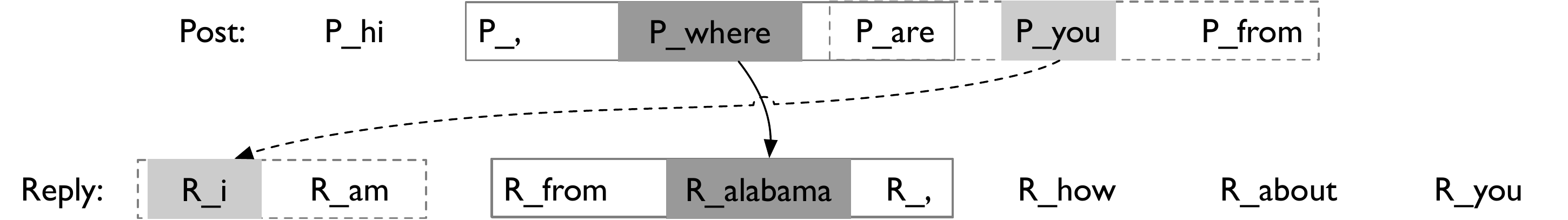}
  \caption{\label{align-show} An example of conversational word alignment from the PersonaChat dataset (section 3.1). `P\_' and `R\_'  identify the vocabulary the words come from. For the word `where,' we find the most related word `alabama' based on the alignment model and generate the cross-sentence window with the size of 3 centered on the word. }
\end {figure*}

\section{Introduction}
Word embedding is one of the most fundamental work in the NLP tasks, where low-dimensional word representations are learned from unlabeled corpora.
The pre-trained embeddings can reflect the semantic and syntactic information of words and help various downstream tasks get better performance \cite{collobert-2011-natural, kim-2014}. 
 
The traditional word embedding methods train the models based on the co-occurrence statistics, such as Word2vec \cite{mikolov-2013-efficient, mikolov-2013-distributed}, GloVe \cite{pennington-2014-glove}.
Those methods are widely used in dialog systems, not only in retrieval-based methods \cite{wang-2015-syntax,Yan-2016-Learning} but also the generation-based models \cite{serban-2016-building,zhang-2018-context}.
The retrieval-based methods predict the answer based on the similarity of context and candidate responses,  which can be divided into single-turn models \cite {wang-2015-syntax} and multi-turn models  \cite{Wu-2017-Sequential,zhou-2018-multi,ma-etal-2019-triplenet} based on the number of turns in context.
Those methods construct the representations of the context and response with a single vector space. Consequently, the models  tend to select the response with the same words .

On the other hand, as those static embeddings can not cope with the phenomenon of polysemy, researchers pay more attention to contextual representations recently.
ELMo \cite{peters-elmo-2018}, BERT \cite {devlin-2019-bert}, and XLNet \cite{yang-2019-xlnet}  have achieved great success in many NLP tasks.
However, it is difficult to apply them in the industrial dialog system due to their low computational efficiency.

In this paper, we focus on the static embedding, for it is flexible and efficient.
The previous works learn the embedding from intra-sentence within a single space, which is not enough for dialog systems. 
Specifically, the semantic correlation beyond a single sentence in the conversation pair is missing. 
For example, the words `why' and `because' usually come from different speakers, and we can not catch their relationship by context window within the sentence.
Furthermore, when the words in post and reply are mapped into the same vector space, the model tends to select boring replies with repeated content because repeated words can easily get a high similarity.

To tackle this problem, we propose PR-Embedding (Post-Reply Embedding) to learn representations from the conversation pairs in different spaces. 
Firstly, we represent the post and the reply in two different spaces similar to the source and target languages in the machine translation.
Then, the word alignment model is introduced to generate the cross-sentence window.
Lastly, we train the embeddings based on the word-level co-occurrence and a sentence-level classification task.

The main contributions of our work are: (1) we propose a new method to learn the conversational word embedding from human dialogue in two different vector spaces; 
(2) The experimental results show that PR-Embedding can help the model select better responses and catch the semantic correlation among the conversation pair.

\section{Methods} 
\subsection{Notation}
We consider two vocabularies for the post and the reply $V^p:=\{v^p_1, v^p_2,...,v^p_s\}, V^r:=\{v^r_1, v^r_2,...,v^r_s\} $ together with two embedding matrices $E_p , E_r \in \mathbb{R}^{s \times d}$, where $s$ is the size of the vocabularity and $d$ is the embedding dimension. 
We need to learn the embedding from the conversation pair  $ \left\langle{post, reply} \right\rangle$. 
They can be formulated as $P = (p_1,...,p_m)$, $R = (r_1,...,r_n)$, where $m, n$ are the length of the post and the reply respectively.
For each pair in the conversation, we represent the post, reply in two spaces $E_p, E_r$, by which we can encode the relationship between the post and reply into the word embeddings.

\subsection{Conversational Word Alignment}
Similar to the previous works \cite{mikolov-2013-distributed, pennington-2014-glove}, we also learn the embeddings based on word co-occurrence.
The difference is that we capture both intra-sentence and cross-sentence co-occurrence.
For the single sentence, the adjacent words usually have a more explicit semantic relation. 
So we also calculate the co-occurrence based on the context window in a fixed size.

However, the relationship among the cross-sentence words is no longer related to their distance.
As shown in Figure \ref {align-show}, the last word in the post `from' is adjacent to the first word `i' in reply, but they have no apparent semantic relation.
So we need to find the most related word from the other sequence for each word in the pair.
In other words, we need to build conversational word alignment between the post and the reply.

In this paper, we solve it by the word alignment model in statistical machine translation \cite{och03-asc-2003}.
We treat the post as the source language and the reply as the target language. 
Then we align the words in the pair with the word alignment model and generate a cross-sentence window centered on the alignment word.

\subsection{Embedding Learning}
We train the conversational word embedding on word and sentence level.

{\noindent{\bf Word-level.} PR-Embedding learns the word representations from the word-level co-occurrence at first.
Following the previous work \cite{pennington-2014-glove}, we train the embedding by the global log-bilinear regression model
\begin {equation}
w_i^T \tilde{w}_k + b_i + \tilde{b}_k = log(X_{ik})
\end{equation}
where $X_{ik}$ is the number of times word $k$ occurs in the context of word $i$. $w$, $ \tilde{w}$ are the word vector and context word vector, $b$ is the bias. 
We construct the word representations by the summation of $w$ and $\tilde{w}$.

{\noindent{\bf Sentence-level.} 
To learn the relationship of embeddings from the two spaces, we further train the embedding by a sentence-level classification task.

We match the words in the post and reply based on the embeddings from word-level learning.
Then we encode the match features by CNN \cite {kim-2014} followed by max-pooling for prediction.
We can formulate it by
\begin {gather}
M_{(i, j)} = cosine(p_i, r_j)  \\
\tilde {M}_{i} = tanh(W_1 \cdot M_{i:i+h-1} + b_1) \\ 
\tilde {M} =  MaxPooling_{i=1}^{m-h+1}[\tilde {M}_{i} ]
\end {gather}
where $W_1, b_1$ are trainable parameters, $M_{i:i+h-1}$ refers to the concatenation of ($M_i, ..., M_{i+h-1}$) and $h$ is the window size of the filter.
At last, we feed the vector $\tilde {M}$ into a fully-connected layer with sigmoid output activation.
\begin {equation}
g(P, R) = sigmoid(W_2 \cdot \tilde {M} + b_2)
\end{equation}
where $W_2, b_2$ are trainable weights. 
We minimize the cross-entropy loss between the prediction and ground truth for training. 

 \begin{table}[t]
        \begin{center}\small
        \begin{tabular}{l ccc}
        \toprule
        & hits@1& hits@5& hits@10  \\ 
        \midrule
        GloVe$_{train}$  & 12.6 & 39.6 & 63.7 \\
        GloVe$_{emb}$ & 18.0 & 44.6 & 66.9 \\
        BERT$_{emb}$ & 15.4 & 41.0 & 62.9 \\
        Fasttext$_{emb}$ & 17.8 & 44.9 & 67.2 \\
        \midrule
        PR-Embedding &\bf 22.4 &\bf 60.0 &\bf 81.1 \\
         \midrule
         \midrule
         IR baseline\dag & 21.4 & - & - \\
         Starpace\dag & 31.8  & - & - \\
         Profile Memory\dag & 31.8  & - & - \\
         \midrule
         KVMemnn  & 32.3 & 62.0 & 79.2 \\
              \quad +PR-Embedding & 35.9 & 66.1 & 82.6 \\
         KVMemnn (GloVe) & 36.8 & 68.1 & 83.6 \\
               \quad +PR-Embedding &\bf  39.9 &\bf 72.4 &\bf 87.0 \\
        \bottomrule
        \end{tabular}
        \end{center}
        \caption{\label{result-persona} Experimental results on the test set of the PersonaChat dataset. The upper part compares the embeddings in the single-turn and the lower one is for the multi-turn task. $train$: train the embedding with the training set; $emb$: use the public embedding directly; \dag: take the results from the paper of the dataset.}
        \end{table}

\section{Experiment} 
\subsection{Datasets}
To better evaluate the embeddings, we choose the manual annotation conversation datasets.
For the English dataset, we use the multi-turn conversation dataset PersonaChat  \cite {zhang-2018-personalizing}.
For the Chinese dataset, we use an in-house labeled test set of the single-turn conversations, which contains 935 posts, and 12767 candidate replies.
Each of the replies has one of the three labels: bad, middle, and good.
The training set comes from Baidu Zhidao \footnote {https://zhidao.baidu.com/} and contains 1.07 million pairs after cleaning.

\subsection{Evaluation}
{\noindent{\bf Baselines.} 
We use GloVe as our main baseline, and compare PR-Embedding with the embedding layer of BERT, which can also be used as static word embedding. 
We also compare with the the public embeddings of Fasttext \cite {joulin-2017-bag} and  DSG \cite {song-2018-directional}.

{\noindent{\bf Tasks.} 
We focus on the response selection tasks for retrieval-based dialogue systems both in the single-turn and multi-turn conversations.
For the Personchat dataset,  we use the current query for response selection in the single-turn task and conduct the experiments in no-persona track because we focus on the relationship between post and reply.

{\noindent{\bf Models.} 
For the single-turn task, we compare the embeddings based on BOW (bag-of-words, the average of all word embedding vectors), and select replies by cosine similarity;
For the multi-turn task, we use a neural model called key-value (KV) memory network \footnote {The official baseline result is 34.9 on hits@1, which is subject to the changes of the computing device.} \cite{miller-etal-2016-key}, which has been proved to be a strong baseline in the ConvAI2 competition \cite{dinan-2020-second}.

{\noindent{\bf Metrics.} 
We use the recall at position $k$ from 20 candidates (hits@k, only one candidate reply is true) as the metrics in the PersonaChat dataset following the previous work \cite {zhang-2018-personalizing}. 
For the Chinese dataset, we use NDCG and P@1 to evaluate the sorted quality of the candidate replies.

{\noindent{\bf Setup.} 
We train the model by Adagrad \cite{duchi-2011-adaptive} and implement it by Keras \cite {chollet-2015-keras} with Tensorflow backend.
For the PersonaChat dataset, we train the embeddings by the training set containing about 10k conversation pairs,  use validation sets to select the best embeddings, and report the performance on test sets.

  \begin{table}[t]
       \begin{center}\small
        \begin{tabular}{l ccccc}
        \toprule
        & NDCG & NDCG@5 & P@1 & P@1(s) \\ 
        \midrule
        GloVe$_{train}$ & 69.97 & 48.87 & 51.23 & 33.48 \\
        DSG$_{emb}$ & 70.82 & 50.45 & 52.19 & 35.61 \\
        BERT$_{emb}$ & 70.06 & 48.45 & 51.66 & 35.08 \\
        \midrule
        PR-Emb & \bf 74.79 &\bf 58.16 &\bf 62.03 & \bf 45.99 \\
        \midrule
           \quad w/o PR & 70.68 & 50.60 & 50.48 & 35.19  \\
           \quad w/o SLL  & 71.65  &  52.03 & 53.48 & 40.86  \\  
        \bottomrule
        \end{tabular}
        \end{center}
        \caption{\label{result-in-house} Experimental results on the Chinese test set. P@1(s) means only use the response with label `good' as the right one and other metrics treat the label `middle' and `good' as right.}
        \end{table}
        
\begin{table*}
\begin{center}
{
\resizebox{\textwidth}{!}{
\begin{tabular}{ccc|ccc|ccc}
\multicolumn{3}{c|}{\textsc{Why}} & \multicolumn{3}{c|}{\textsc{Thanks}} & \multicolumn{3}{c}{\textsc{Congratulations}} \\ \hline 
\multicolumn{1}{c}{GloVe} & \multicolumn{1}{c}{P-Emb}  & \multicolumn{1}{c|}{R-Emb} & \multicolumn{1}{c}{GloVe} & \multicolumn{1}{c}{P-Emb} & \multicolumn{1}{c|}{R-Emb}  &\multicolumn{1}{c}{GloVe} & \multicolumn{1}{c}{P-Emb}  & \multicolumn{1}{c}{R-Emb} \\ 
\hline\hline
why       & why                & because            & thanks    & thanks   & welcome        & congratulations  & congratulations    & thank  \\
know     & understand     & matter            & thank       & asking    & problem         & congrats        & ah                      & thanks \\ 
guess    & oh                  & idea               & fine        & thank       & today             & goodness     & fantastic             & appreciate \\
so          & probably       & reason          & asking     & good       &bill                & yum               & bet                      & problem  \\
\hline
\end{tabular}
}
}
\end{center}
\caption{\label{closestpairs} Four nearest tokens for the selected words trained by our PR-Embedding (P/R-Emb) and GloVe.  }
\end{table*}

\subsection{Results}
The results on the PersonaChat dataset are shown in Table \ref {result-persona}.
The strongest baseline in the single-turn task is GloVe, but PR-Embedding outperforms the baseline by 4.4\%.
For the multi-turn task, we concatenate PR-Embeddings with the original embedding layer of the model.
We find that the performance becomes much better when we concatenate PR-Embedding with the randomly initialized embedding.
The model KVMemnn becomes much stronger when the embedding layer initializes with the embeddings from GloVe.
However, PR-Embedding still improves the performance significantly.

The results on the in-house dataset are in Table \ref {result-in-house}.
Our method (PR-Emb) significantly exceeds all the baselines in all metrics.
The improvement is greater than the results on the English dataset as the training corpus is much larger.
Note that, all the improvements on both datasets are statistically significant (p-value $\leq 0.01$).

\subsection{Ablation}
We conduct the ablations on Chinese datasets in consideration of its larger training corpus. The results are in the last part of Table \ref {result-in-house}.
When we change the two vector spaces into the single one (w/o PR), the model is similar to GloVe with sentence-level learning.
The performance becomes much worse in all the metrics, which shows the effect of two vector spaces.
Furthermore, all the scores drop significantly after sentence-level learning is removed (w/o SLL), which shows its necessity.

\section{Analysis} 
\subsection{Nearest  Tokens}
We provide an analysis based on the nearest tokens for the selected words in the whole vector space, including the word itself.
For PR-Embedding, we select the words from the post vocabulary and give the nearest words both in the post and the reply space.
Note that all of them are trained by the training set of the PersonaChat dataset.

The results are in Table \ref{closestpairs}. 
For the columns in GloVe and P-Emb, the words are the same (first one) or similar to the selected ones because the nearest token for any word is itself within a single vector space.
The similarity makes that the model tends to select the reply with repeated words.
While the words in the column R-Emb are relevant to the selected words, such as words  `why' and `because,' `thanks' and `welcome,' `congratulations' and `thank.' 
Those pairs indicate that PR-Embedding catches the correlation among the conversation pairs, which is helpful for the model to select the relevant and content-rich reply.

\begin {figure} [t]
  \centering
  \includegraphics [width= 0.48\textwidth]  {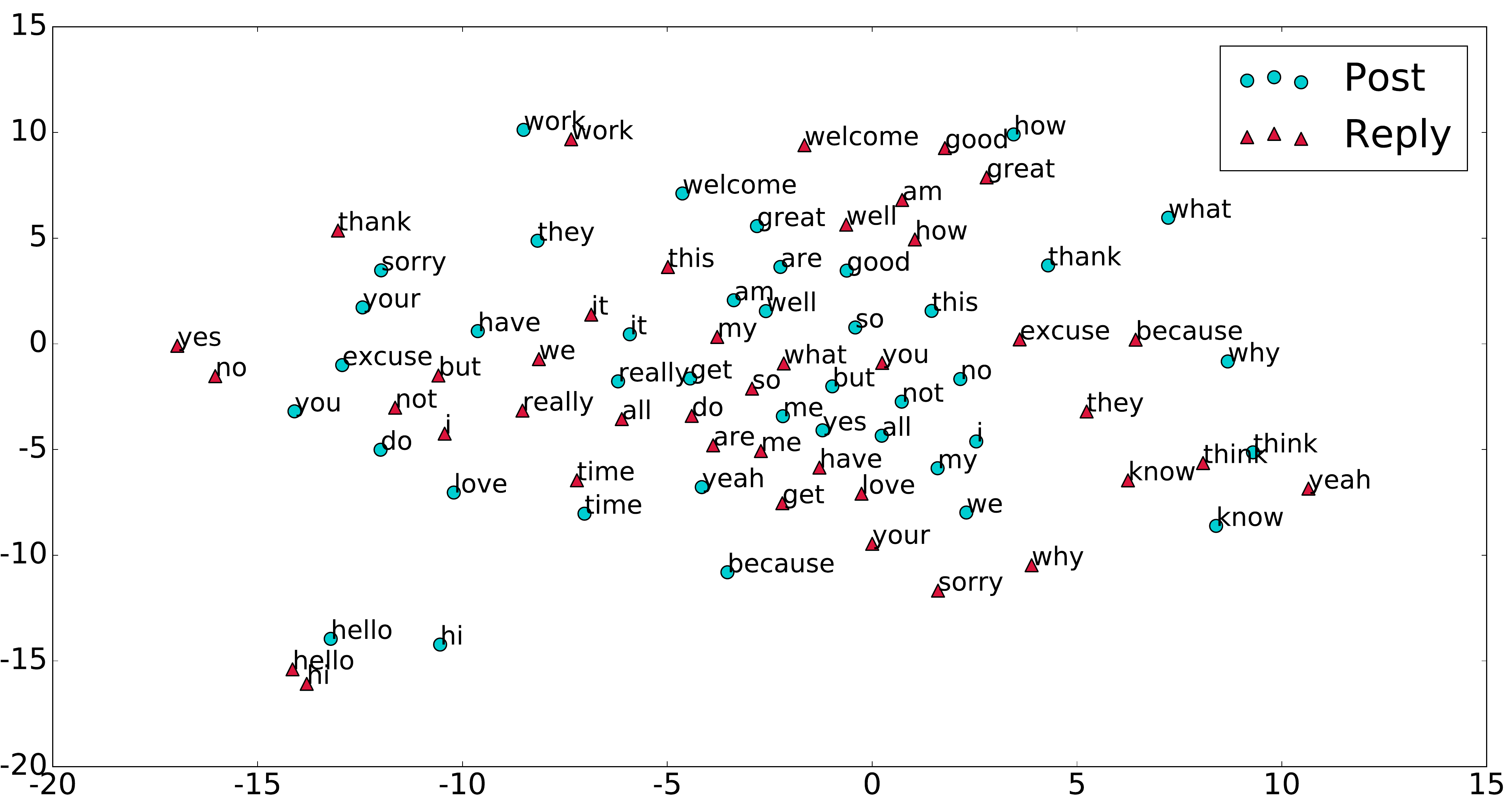}
  \caption{\label{visual-pr} The visualization of the 40 words with the highest frequency in PR-Embedding.}
\end {figure}

\subsection{Visualization}
To further explore how PR-Embedding represents words and the relation between the two spaces,  we use t-SNE \cite {maaten-2008-visualizing} to visualize the embeddings of  40 words with the highest frequency except for stop words in the spaces.  

The embeddings are visualized in Figure \ref {visual-pr}.
For the embeddings in the same spaces, the words with similar semantic meanings are close to each other, indicating that PR-Embedding catches the similarity within the same space.
For example, the words `hello' and `hi', `good' and `great', `not' and `no'.

For the same words in different spaces, most of them have close locations, especially nouns and verbs, such as `work,' `think,' `know.' 
Maybe it is because they play a similar role in the post and the reply.
While some question words have different situations, for example,  `how' and `good, great,' `why' and `because' show the clear relations in the post and the reply spaces, which conforms to the habit of human dialog. 
Furthermore, PR-Embeddings can also capture the correlation between pronouns such as such as `my, we' and `your' also catch the correlation.
We can conclude that our method can encode the correlation among the two spaces into the embeddings. 

\section{Conclusions} 
In this paper, we have proposed a conversational word embedding method named PR-Embedding, which is learned from conversation pairs for retrieval-based dialog system.
We use the word alignment model from machine translation to calculate the cross-sentence co-occurrence and train the embedding on word and sentence level.
We find that PR-Embedding can help the models select the better response both in single-turn and multi-turn conversation by catching the information among the pairs.
In the future, we will adapt the method to more neural models especially the generation-based methods for the dialog system.

 \section*{Acknowledgement}\label{Acknowledgement}
 We would like to thank all anonymous reviewers for their hard work on reviewing and providing valuable comments on our paper.  We thank Yecheng Hu and Chenglei Si for proofreading our paper thoroughly. We also would like to thank Quan Wen for insightful suggestions.  This work was supported by the National Natural Science Foundation of China (NSFC) via grant 61976072, 61632011 and 61772153.

\bibliography{anthology,acl2020}

\begin{thebibliography}{24}
\expandafter\ifx\csname natexlab\endcsname\relax\def\natexlab#1{#1}\fi

\bibitem[{Chollet et~al.(2015)}]{chollet-2015-keras}
Fran\c{c}ois Chollet et~al. 2015.
\newblock Keras.
\newblock \url{https://keras.io}.

\bibitem[{Collobert et~al.(2011)Collobert, Weston, Bottou, Karlen, Kavukcuoglu,
  and Kuksa}]{collobert-2011-natural}
Ronan Collobert, Jason Weston, L{\'e}on Bottou, Michael Karlen, Koray
  Kavukcuoglu, and Pavel Kuksa. 2011.
\newblock Natural language processing (almost) from scratch.
\newblock \emph{Journal of machine learning research}, 12(Aug):2493--2537.

\bibitem[{Devlin et~al.(2019)Devlin, Chang, Lee, and
  Toutanova}]{devlin-2019-bert}
Jacob Devlin, Ming-Wei Chang, Kenton Lee, and Kristina Toutanova. 2019.
\newblock Bert: Pre-training of deep bidirectional transformers for language
  understanding.
\newblock In \emph{Proceedings of the 2019 Conference of the North American
  Chapter of the Association for Computational Linguistics: Human Language
  Technologies, Volume 1 (Long and Short Papers)}, pages 4171--4186.

\bibitem[{Dinan et~al.(2020)Dinan, Logacheva, Malykh, Miller, Shuster, Urbanek,
  Kiela, Szlam, Serban, Lowe et~al.}]{dinan-2020-second}
Emily Dinan, Varvara Logacheva, Valentin Malykh, Alexander Miller, Kurt
  Shuster, Jack Urbanek, Douwe Kiela, Arthur Szlam, Iulian Serban, Ryan Lowe,
  et~al. 2020.
\newblock The second conversational intelligence challenge (convai2).
\newblock In \emph{The NeurIPS'18 Competition}, pages 187--208. Springer.

\bibitem[{Duchi et~al.(2011)Duchi, Hazan, and Singer}]{duchi-2011-adaptive}
John Duchi, Elad Hazan, and Yoram Singer. 2011.
\newblock Adaptive subgradient methods for online learning and stochastic
  optimization.
\newblock \emph{Journal of Machine Learning Research}, 12(Jul):2121--2159.

\bibitem[{Joulin et~al.(2017)Joulin, Grave, Bojanowski, and
  Mikolov}]{joulin-2017-bag}
Armand Joulin, Edouard Grave, Piotr Bojanowski, and Tomas Mikolov. 2017.
\newblock \href {https://www.aclweb.org/anthology/E17-2068} {Bag of tricks for
  efficient text classification}.
\newblock In \emph{Proceedings of the 15th Conference of the {E}uropean Chapter
  of the Association for Computational Linguistics: Volume 2, Short Papers},
  pages 427--431, Valencia, Spain. Association for Computational Linguistics.

\bibitem[{Kim(2014)}]{kim-2014}
Yoon Kim. 2014.
\newblock \href {https://doi.org/10.3115/v1/D14-1181} {Convolutional neural
  networks for sentence classification}.
\newblock In \emph{Proceedings of EMNLP 2014}, pages 1746--1751. Association
  for Computational Linguistics.

\bibitem[{Ma et~al.(2019)Ma, Cui, Shao, He, Zhang, Liu, Wang, and
  Hu}]{ma-etal-2019-triplenet}
Wentao Ma, Yiming Cui, Nan Shao, Su~He, Wei-Nan Zhang, Ting Liu, Shijin Wang,
  and Guoping Hu. 2019.
\newblock \href {https://www.aclweb.org/anthology/K19-1069} {{T}riple{N}et:
  Triple attention network for multi-turn response selection in retrieval-based
  chatbots}.
\newblock In \emph{Proceedings of the 23rd Conference on Computational Natural
  Language Learning (CoNLL)}, pages 737--746, Hong Kong, China. Association for
  Computational Linguistics.

\bibitem[{Maaten and Hinton(2008)}]{maaten-2008-visualizing}
Laurens van~der Maaten and Geoffrey Hinton. 2008.
\newblock Visualizing data using t-sne.
\newblock \emph{Journal of machine learning research}, 9(Nov):2579--2605.

\bibitem[{Mikolov et~al.(2013{\natexlab{a}})Mikolov, Chen, Corrado, and
  Dean}]{mikolov-2013-efficient}
Tomas Mikolov, Kai Chen, Greg Corrado, and Jeffrey Dean. 2013{\natexlab{a}}.
\newblock Efficient estimation of word representations in vector space.
\newblock \emph{arXiv preprint arXiv:1301.3781}.

\bibitem[{Mikolov et~al.(2013{\natexlab{b}})Mikolov, Sutskever, Chen, Corrado,
  and Dean}]{mikolov-2013-distributed}
Tomas Mikolov, Ilya Sutskever, Kai Chen, Greg~S Corrado, and Jeff Dean.
  2013{\natexlab{b}}.
\newblock Distributed representations of words and phrases and their
  compositionality.
\newblock In \emph{Advances in neural information processing systems}, pages
  3111--3119.

\bibitem[{Miller et~al.(2016)Miller, Fisch, Dodge, Karimi, Bordes, and
  Weston}]{miller-etal-2016-key}
Alexander Miller, Adam Fisch, Jesse Dodge, Amir-Hossein Karimi, Antoine Bordes,
  and Jason Weston. 2016.
\newblock \href {https://doi.org/10.18653/v1/D16-1147} {Key-value memory
  networks for directly reading documents}.
\newblock In \emph{Proceedings of the 2016 Conference on Empirical Methods in
  Natural Language Processing}, pages 1400--1409, Austin, Texas. Association
  for Computational Linguistics.

\bibitem[{Och and Ney(2003)}]{och03-asc-2003}
Franz~Josef Och and Hermann Ney. 2003.
\newblock A systematic comparison of various statistical alignment models.
\newblock \emph{Computational Linguistics}, 29(1):19--51.

\bibitem[{Pennington et~al.(2014)Pennington, Socher, and
  Manning}]{pennington-2014-glove}
Jeffrey Pennington, Richard Socher, and Christopher Manning. 2014.
\newblock Glove: Global vectors for word representation.
\newblock In \emph{Proceedings of the 2014 conference on empirical methods in
  natural language processing (EMNLP)}, pages 1532--1543.

\bibitem[{Peters et~al.(2018)Peters, Neumann, Iyyer, Gardner, Clark, Lee, and
  Zettlemoyer}]{peters-elmo-2018}
Matthew Peters, Mark Neumann, Mohit Iyyer, Matt Gardner, Christopher Clark,
  Kenton Lee, and Luke Zettlemoyer. 2018.
\newblock \href {https://doi.org/10.18653/v1/N18-1202} {Deep contextualized
  word representations}.
\newblock In \emph{Proceedings of NAACL-HLT-2018}, pages 2227--2237.
  Association for Computational Linguistics.

\bibitem[{Serban et~al.(2016)Serban, Sordoni, Bengio, Courville, and
  Pineau}]{serban-2016-building}
Iulian~V Serban, Alessandro Sordoni, Yoshua Bengio, Aaron Courville, and Joelle
  Pineau. 2016.
\newblock Building end-to-end dialogue systems using generative hierarchical
  neural network models.
\newblock In \emph{Proceedings of the Thirtieth AAAI Conference on Artificial
  Intelligence}, pages 3776--3783.

\bibitem[{Song et~al.(2018)Song, Shi, Li, and Zhang}]{song-2018-directional}
Yan Song, Shuming Shi, Jing Li, and Haisong Zhang. 2018.
\newblock Directional skip-gram: Explicitly distinguishing left and right
  context for word embeddings.
\newblock In \emph{Proceedings of the 2018 Conference of the North American
  Chapter of the Association for Computational Linguistics: Human Language
  Technologies, Volume 2 (Short Papers)}, pages 175--180.

\bibitem[{Wang et~al.(2015)Wang, Lu, Li, and Liu}]{wang-2015-syntax}
Mingxuan Wang, Zhengdong Lu, Hang Li, and Qun Liu. 2015.
\newblock Syntax-based deep matching of short texts.
\newblock In \emph{Proceedings of the 24th International Conference on
  Artificial Intelligence}, pages 1354--1361. AAAI Press.

\bibitem[{Wu et~al.(2017)Wu, Wu, Xing, Zhou, and Li}]{Wu-2017-Sequential}
Yu~Wu, Wei Wu, Chen Xing, Ming Zhou, and Zhoujun Li. 2017.
\newblock \href {https://doi.org/10.18653/v1/P17-1046} {Sequential matching
  network: A new architecture for multi-turn response selection in
  retrieval-based chatbots}.
\newblock In \emph{Proceedings of ACL 2017}, pages 496--505. Association for
  Computational Linguistics.

\bibitem[{Yan et~al.(2016)Yan, Song, and Wu}]{Yan-2016-Learning}
Rui Yan, Yiping Song, and Hua Wu. 2016.
\newblock Learning to respond with deep neural networks for retrieval-based
  human-computer conversation system.
\newblock In \emph{International ACM SIGIR Conference on Research and
  Development in Information Retrieval}, pages 55--64.

\bibitem[{Yang et~al.(2019)Yang, Dai, Yang, Carbonell, Salakhutdinov, and
  Le}]{yang-2019-xlnet}
Zhilin Yang, Zihang Dai, Yiming Yang, Jaime Carbonell, Ruslan Salakhutdinov,
  and Quoc~V Le. 2019.
\newblock Xlnet: Generalized autoregressive pretraining for language
  understanding.
\newblock \emph{arXiv preprint arXiv:1906.08237}.

\bibitem[{Zhang et~al.(2018{\natexlab{a}})Zhang, Dinan, Urbanek, Szlam, Kiela,
  and Weston}]{zhang-2018-personalizing}
Saizheng Zhang, Emily Dinan, Jack Urbanek, Arthur Szlam, Douwe Kiela, and Jason
  Weston. 2018{\natexlab{a}}.
\newblock Personalizing dialogue agents: I have a dog, do you have pets too?
\newblock In \emph{Proceedings of the 56th Annual Meeting of the Association
  for Computational Linguistics (Volume 1: Long Papers)}, pages 2204--2213.

\bibitem[{Zhang et~al.(2018{\natexlab{b}})Zhang, Cui, Wang, Zhu, Li, Zhou, and
  Liu}]{zhang-2018-context}
Weinan Zhang, Yiming Cui, Yifa Wang, Qingfu Zhu, Lingzhi Li, Lianqiang Zhou,
  and Ting Liu. 2018{\natexlab{b}}.
\newblock Context-sensitive generation of open-domain conversational responses.
\newblock In \emph{Proceedings of the 27th International Conference on
  Computational Linguistics}, pages 2437--2447.

\bibitem[{Zhou et~al.(2018)Zhou, Li, Dong, Liu, Chen, Zhao, Yu, and
  Wu}]{zhou-2018-multi}
Xiangyang Zhou, Lu~Li, Daxiang Dong, Yi~Liu, Ying Chen, Wayne~Xin Zhao, Dianhai
  Yu, and Hua Wu. 2018.
\newblock Multi-turn response selection for chatbots with deep attention
  matching network.
\newblock In \emph{Proceedings of ACL 2018}, volume~1, pages 1118--1127.

\end{thebibliography}
\bibliographystyle{acl_natbib}
\end{document}